%% file: EDM_Article_Submission.tex
\documentclass{edm_article}

\usepackage{algorithm2e}
\usepackage{xcolor}         
\usepackage{changepage} 

\newcommand{\comment}[1]{}

\usepackage{booktabs, tabularx}

\begin{document}

\title{Increasing Students' Engagement to Reminder Emails Through Multi-Armed Bandits}
%
\numberofauthors{6}
\author{
Fernando J. Yanez\\
       \affaddr{University of Toronto}\\
       \email{fyanez@cs.toronto.edu}
\and
Angela Zavaleta-Bernuy\\
       \affaddr{University of Toronto}\\
       \email{angelazb@cs.toronto.edu}
\and
Ziwen Han\\
       \affaddr{University of Toronto}\\
       \email{ziwen.han@mail.utoronto.ca}
\and
Michael Liut\\
       \affaddr{University of Toronto}\\
       \email{michael.liut@utoronto.ca}
\and
Anna Rafferty\\
       \affaddr{Carleton College}\\
       \email{arafferty@carleton.edu}
\and
Joseph Jay Williams\\
       \affaddr{University of Toronto}\\
       \email{williams@cs.toronto.edu}
}

\maketitle

\input{sections/0_abstract}
\input{sections/1_introduction}

\input{sections/2_design}

\input{sections/4_methods}
\input{sections/5_results}
\input{sections/6_discussion}
\input{sections/7_conclusion}

\bibliographystyle{abbrv}
\bibliography{sigproc}  

\end{document}

%% file: sections/0_abstract.tex
\begin{abstract}
Conducting randomized experiments in education settings raises the question of how we can use machine learning techniques to improve educational interventions. Using Multi-Armed Bandits (MAB) algorithms like Thompson Sampling (TS) in adaptive experiments can increase students' chances of obtaining better outcomes by increasing the probability of assignment to the most optimal condition (arm), even before an intervention completes. This is an advantage over traditional A/B testing, which may allocate an equal number of students to both optimal and non-optimal conditions. The problem is the exploration-exploitation trade-off. Even though adaptive policies aim to collect enough information to allocate more students to better arms reliably, past work shows that this may not be enough exploration to draw reliable conclusions about whether arms differ. Hence, it is of interest to provide additional uniform random (UR) exploration throughout the experiment. This paper shows a real-world adaptive experiment on how students engage with instructors' weekly email reminders to build their time management habits. Our metric of interest is open email rates which tracks the arms represented by different subject lines. These are delivered following different allocation algorithms: UR, TS, and what we identified as $\text{TS}^\dagger$—which combines both TS and UR rewards to update its priors. We highlight problems with these adaptive algorithms—such as possible exploitation of an arm when there is no significant difference—and address their causes and consequences. Future directions includes studying situations where the early choice of the optimal arm is not ideal and how adaptive algorithms can address them. 


\keywords{Multi-armed bandits, Randomized experiments, A/B testing, Field deployment, Reinforcement learning.}\\
\linebreak
\end{abstract}

%% file: sections/1_introduction.tex
\section{Introduction}

Traditional A/B randomized comparisons are widely used in different research and industry areas to determine if a particular condition is more effective. Similarly, adaptive experimentation is used to assign participants to the most effective current condition while keeping the ability to test the other conditions \cite{williams2018enhancing}. Using adaptive experimentation in education can help explore various conditions but also direct more students to more useful ones in a randomized experiment \cite{lomas2016interface, williams2016axis, williams2018enhancing}. 



In this setting, we focus on the adaptive experimentation's problem regarding the exploration-exploitation trade-off. It is well studied that some adaptive policies might be exploiting an arm when no significant difference exists in that direction \cite{nogas2021algorithms, williams2016axis}. This might cause a lack of exploration of all conditions where the policy starts allocating more users to an arm that was favoured from high values of its random rewards, even if the true mean is higher. We study the impact of adding a fixed allocation of uniform exploration to the TS algorithm—which is employed to solve the MABs problem and has been used in A/B experiments to adaptively assign participants to conditions \cite{williams2018enhancing}. TS is a Bayesian algorithm that has shown promising results in maximizing the reward \cite{agrawal2012analysis, chapelle2011empirical}. It maintains a posterior distribution for each action based on how rewarding that action is expected to be. Actions are chosen proportional to their posterior probability, and the outcome of the chosen action is then used to update each posterior distribution. TS also performs well in cases where batches of actions are chosen, and the rewards and outcomes are updated only after each batch \cite{kalkanli2021asymptotic}.


This work aims to build onto the study of academic interventions targeted to enhance study habits and time management of college students, building upon \cite{zavaleta2021using} by studying the variation from injecting UR data to TS priors each week. Previous work focused on some methods that improve student's study habits and attitudes \cite{aquino2011study},
and using A/B comparisons in email interventions to students \cite{zavaleta2022can}. 

We conducted a real-world adaptive experiment in an online synchronous CS1 course at a large publicly funded research-focused North American University. The motivation for this intervention is to show students that simply knowing how to program is only one part of the puzzle; building consistent study habits and depositions such as time management are also crucial to being a good programmer. These skills aim to impact longer-term success, in addition to capturing early improvement from keeping up with course content. 

We explore the difference between implementing a smaller traditional experiment (i.e., allocating students to conditions uniformly at random) in addition to an adaptive experiment (i.e., TS) mixed on the same cohort. This approach is motivated by creating stability in the adaptive component and increasing the probability of assigning more students to a better condition. Even more, it could make it easier to draw conclusions from the experiment since we can look at the traditional part of the experiment or the combined one. In total, we deployed three policy learning algorithms. The first is simple uniform random (UR) sampling, which is assigned to half the cohort each week. The second one is a standard TS algorithm that was introduced after a burn-in period of four weeks. Lastly, the third group is introduced a week after the TS algorithm and uses a combination of uniformly sampled and TS sampled evidence to update its beliefs each week, which we refer to as $\text{TS}^\dagger$. This allows us to evaluate adaptive methods (TS) and uniform random side by side and evaluate how a combined version of these algorithms behaves. A more thorough explanation can be found in \S\ref{sec:design} and \S\ref{sec:methods}.

%% file: sections/2_design.tex
\section{Real-world Intervention}
\label{sec:design}
We conducted our reminder email intervention in a CS1 course lasting 13 weeks. Students were randomly assigned to a subject line condition each week. The messages were sent on the same day and time every week (i.e. Wednesday evenings). There were initially $1119$ students enrolled in the course, with a final count of $1025$ by the end of the term.

These emails are aimed at the goal of building time management skills for students, we sent weekly reminder emails with three different versions of subject lines. The email content nudged students to start their weekly homework early. Our end goal is to find better strategies to motivate students to interact with the email content, which we cannot do if they do not open these emails. Hence, we wanted to determine whether one subject line would captivate students more, thereby motivating them to engage with our prompts.

As it is difficult to fully observe student engagement to asynchronous emails, we chose to collect data on the open rates of the messages as our response measurement of engagement in this intervention. To assign subject lines, we used variations of adaptive multi-armed bandit algorithms to give one of three different versions of the email subject line each week. Data from the previous week is then used to further allocate email subject lines for the coming week, as described in Section 3. This allows us to investigate which might be more effective in leading students to open the reminder email and pivot towards sending this subject line more under the assumption its impact is identical from week to week.  

To design the subject lines, we used various psychological and marketing theories.
\begin{adjustwidth}{-1em}{}
\begin{enumerate}
    \item[] \textbf{Subject Line 1 (SL1)}: \textit{Hey \{name\}, when are you going to make a plan for your {{course code}} homework assignments?} This subject line was designed using implementation intentions to help students transform the goal of planning when to start their homework into an action \cite{gollwitzer1999implementation}. Using the student’s first-name shows email personalization and can catch their attention easily \cite{sahni_wheeler_chintagunta_2018}.
    \item[] \textbf{Subject Line 2 (SL2)}: \textit{[{\{course code\}} Action Required] Hey \{name\}, start planning your week now!} \includegraphics[scale=0.8]{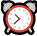}: This subject line adds a sense of urgency to the email, highlighting the importance of planning their homework, and has an emoji at the end to make it stand out in their inbox \cite{valenzuela2022boost}. This subject line also includes the student’s first name \cite{sahni_wheeler_chintagunta_2018}.
    \item[] \textbf{Subject Line 3 (SL3)}: \textit{Reminder: These are your {{course code}} tasks for next week.} This subject line only mentions the content of the email and does not include any personalization or action prompt i.e. does not use any of the theories described above.
\end{enumerate}
\end{adjustwidth}

%% file: sections/4_methods.tex
\section{Methods: Traditional and Adaptive Experimentation}
\label{sec:methods}

To address the exploration versus exploitation problem in MAB settings, we use a set of three algorithms which all share data from a uniform burn-in period of 4 weeks. The uniform algorithm allocation is continued at half of the allocations through all the weeks post burn-in to ensure we are collecting enough data for all arms and to maintain enough samples from all arms. Pure TS—which selects an arm based on its prior Beta distribution with parameters $\alpha$ and $\beta$, representing measures for \textit{successes} and \textit{failures} respectively—accounts for another quarter of the allocations and is used to evaluate the effectiveness and problems of adaptive algorithms. 

We introduce a new algorithm $\text{TS}^\dagger$ for the last quarter of allocations. This uses the same stochastic allocation as TS but draws half-weighted priors from both itself and the uniform allocation to explicitly maintain exploration. We hypothesize that this third allocation algorithm can balance the rapid choice of optimal arms and better adapt in the face of potentially non-stationary arm means.

Given that the adaptive portion of our experiment relies on the Thompson Sampling policy, we explicitly show how the priors $\left(\alpha_{k, t}^{\pi},\ \beta_{k, t}^{\pi}\right)$ change over time, both in equations and with Algorithm \ref{alg:method}. Note in the equations that the priors are updated in a weekly batch. Hence the TS algorithm was chosen for its stochastic nature to avoid assigning all students to the same condition each week. This weekly update ensures time is not an uncontrollable factor affecting students' willingness to open emails.

\subsection{Equations}

Let $k \in \{1,2,3\}$ represent the arm a particular student was assigned at any time $t \in \{1, 2, ..., 13\}$ in weeks. $r_{k,t}^\pi$ represents the binomial distributed cumulative reward from all the students assigned to arm $k$ for policy $\pi$ at time $t$, and $n_{k,t}^\pi$ is the total number of students assigned to arm $k$ for policy $\pi$ at time $t$.

\begin{gather*}
    \left(\alpha_{k, t}^{\text{UR}},\ \beta_{k, t}^{\text{UR}}\right)=\left(1+\sum_{\tau=1}^t\left[r_{k, \tau}^{\text{UR}}\right], 1+\sum_{\tau=1}^t \left[n_{k,\tau}^{\text{UR}}-r_{k, \tau}^{\text{UR}}\right]\right) \\ \text{for } t \leq 13\\
    \\
    \left(\alpha_{k, t}^{\text{TS}},\ \beta_{k, t}^{\text{TS}}\right) = \\
        \begin{cases}
        \left(\alpha_{k, t}^{\text{UR}},\ \beta_{k, t}^{\text{UR}}\right) & \text{for } t \leq 5\\
        \left(\alpha_{k, t-1}^{\text{TS}},\ \beta_{k, t-1}^{\text{TS}}\right)\\ \qquad + \left(r_{k, t-1}^{\text{TS}},\ n_{k, t-1}^{\text{TS}} - r_{k, t-1}^{\text{TS}}\right) & \text{for } 6 \leq t \leq 13\\
        \end{cases}\\
    \\
    \left(\alpha_{k, t}^{\text{TS}^\dagger},\ \beta_{k, t}^{\text{TS}^\dagger}\right) = \\
        \begin{cases}
        \left(\alpha_{k, t}^{\text{UR}},\ \beta_{k, t}^{\text{UR}}\right) & \text{for } t \leq 5\\
        \left(\alpha_{k, t-1}^{\text{TS}^\dagger},\ \beta_{k, t-1}^{\text{TS}^\dagger}\right)\\ \qquad + \frac{1}{2}\left(r_{k, t-1}^{\text{TS}},\ n_{k, t-1}^{\text{TS}} - r_{k, t-1}^{\text{TS}}\right)\\ \qquad + \frac{1}{2}\left(r_{k, t-1}^{\text{UR}},\ n_{k, t-1}^{\text{UR}} - r_{k, t-1}^{\text{UR}}\right)& \text{for } t = 6\\
        \left(\alpha_{k, t-1}^{\text{TS}^\dagger},\ \beta_{k, t-1}^{\text{TS}^\dagger}\right)\\ \qquad + \frac{1}{2}\left(r_{k, t-1}^{\text{TS}^\dagger},\ n_{k, t-1}^{\text{TS}^\dagger} - r_{k, t-1}^{\text{TS}^\dagger}\right)\\ \qquad + \frac{1}{2}\left(r_{k, t-1}^{\text{UR}},\ n_{k, t-1}^{\text{UR}} - r_{k, t-1}^{\text{UR}}\right) & \text{for } 7 \leq t \leq 13\\
        \end{cases}
\end{gather*}

\RestyleAlgo{ruled}
\SetKwComment{Comment}{/* }{ */}
\begin{algorithm}[hbt!]
\caption{Update rules per policy}\label{alg:method}
$\pi \in \{\text{UR}, \text{TS}, \text{TS}^\dagger\}$\\
$k \in \{1,2,3\}$\\
$\left(\alpha_{k, 0}^{\pi},\ \beta_{k, 0}^{\pi}\right) \gets \left(1,\ 1\right), \forall \pi, k$ \Comment*[r]{Initialize parameters}
\For{$t = 1, 2, ..., 5$}{
    $n_{k,t}^\text{UR}$ \Comment*[r]{Allocate with UR}
    $r_{k,t}^\text{UR}$ \Comment*[r]{Get total reward}
    $\left(\alpha_{k, t}^{\pi}, \beta_{k, t}^{\pi}\right) \gets \left(\alpha_{k, t-1}{^{\pi}},\ \beta_{k, t-1}^\pi\right) + \left(r_{k,t}^\text{UR},\ n_{k,t}^\text{UR} - r_{k,t}^\text{UR}\right)$ \Comment*[r]{Update priors}
}
\For{$t = 6$}{
    \For{$\pi' \in \{\text{UR}, \text{TS}\}$}{
        $n_{k,t}^{\pi'}$ \Comment*[r]{Allocate with policy $\pi'$}
        $r_{k,t}^{\pi'}$ \Comment*[r]{Get total reward per $\pi'$}
        $\left(\alpha_{k, t}^{\pi'},\ \beta_{k, t}^{\pi'}\right) \gets \left(\alpha_{k, t-1}^{\pi'},\ \beta_{k, t-1}^{\pi'}\right) + \left(r_{k,t}^{\pi'},\ n_{k,t}^{\pi'} - r_{k,t}^{\pi'}\right)$ \Comment*[r]{Update priors}
    }
    $\left(\alpha_{k, t}^{\text{TS}^\dagger},\ \beta_{k, t}^{\text{TS}^\dagger}\right) \gets \left(\alpha_{k, t-1}^{\text{TS}^\dagger},\ \beta_{k, t-1}^{\text{TS}^\dagger}\right) + \frac{1}{2}\left(r_{k,t}^{\text{TS}},\ n_{k,t}^{\text{TS}} - r_{k,t}^{\text{TS}}\right) + \frac{1}{2}\left(r_{k,t}^{\text{UR}},\ n_{k,t}^{\text{UR}} - r_{k,t}^{\text{UR}}\right)$ \Comment*[r]{Update $\text{TS}^\dagger$ prior}
}
\For{$t = 7, 8, ..., 13$}{
        $n_{k,t}^{\pi}$ \Comment*[r]{Allocate with policy $\pi$}
    $r_{k,t}^{\pi}$ \Comment*[r]{Get total reward per policy $\pi$}

    \eIf{$\pi = \text{TS}^\dagger$}{
        $\left(\alpha_{k, t}^{\text{TS}^\dagger},\ \beta_{k, t}^{\text{TS}^\dagger}\right) \gets \left(\alpha_{k, t-1}^{\text{TS}^\dagger},\ \beta_{k, t-1}^{\text{TS}^\dagger}\right) + \frac{1}{2}\left(r_{k,t}^{\text{TS}^\dagger},\ n_{k,t}^{\text{TS}^\dagger} - r_{k,t}^{\text{TS}^\dagger}\right) + \frac{1}{2}\left(r_{k,t}^{\text{UR}},\ n_{k,t}^{\text{UR}} - r_{k,t}^{\text{UR}}\right)$ \Comment*[r]{Update $\text{TS}^\dagger$ prior}
    }{
        $\left(\alpha_{k, t}^{\pi},\ \beta_{k, t}^{\pi}\right) \gets \left(\alpha_{k, t-1}^{\pi},\ \beta_{k, t-1}^{\pi}\right) + \left(r_{k,t}^{\pi},\ n_{k,t}^{\pi} - r_{k,t}^{\pi}\right)$ \Comment*[r]{Update priors}
    }
}
\end{algorithm}



%% file: sections/5_results.tex
\section{Analysis and Results}
\label{sec:results}
\begin{figure*} \label{fig:experiment}
    \includegraphics[width=\textwidth]{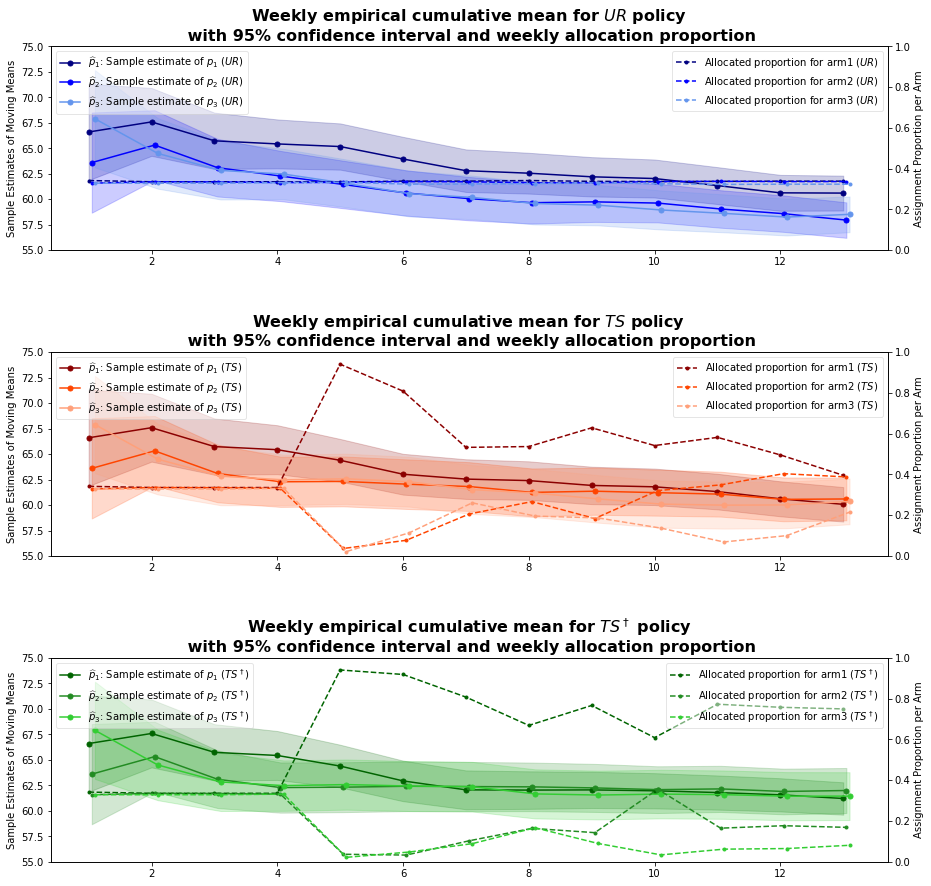}
    \caption{Cumulative sample arm means (hard lines) and weekly allocation proportions (dashed lines) for each policy (blue: UR, red: TS, green: $\text{TS}^\dagger$). Error bar gradients represent 95\% confidence intervals.}
\end{figure*}

Figure \ref{fig:experiment} represents the behaviour of each of the arms in the experiment per allocation algorithm. Each plot represents a different algorithm from UR, TS, $\text{TS}^\dagger$. Note that since all algorithms draw from the same burn-in period, the first 4 weeks are duplicated across plots. For TS and $\text{TS}^\dagger$, an additional week of duplication is shared due to initialization of priors from $\text{TS}^\dagger$.

The left vertical axis represents the cumulative reward mean per week, i.e. the proportion of students who opened the email each week (with 95\% confidence interval in shades), whereas the right vertical axis shows the assignment proportion to every arm week-wise. Furthermore, to understand the output of the experiment in absolute values, Table \ref{tab:empirical_data} displays the final cumulative empirical mean, and the final number of students allocated to each arm throughout the experiment.

\begin{table*}
    \centering
    \begin{tabular}{c|c|c|c|c|c|c|c|c|c}
        \toprule
        & \multicolumn{3}{c|}{UR}
        & \multicolumn{3}{c|}{TS}
        & \multicolumn{3}{c}{$\text{TS}^\dagger$}\\
        Arm
        & \multicolumn{1}{c|}{Mean}
        & \multicolumn{1}{c|}{SD}
        & \multicolumn{1}{c|}{Observations}
        & \multicolumn{1}{c|}{Mean}
        & \multicolumn{1}{c|}{SD}
        & \multicolumn{1}{c|}{Observations}
        & \multicolumn{1}{c|}{Mean}
        & \multicolumn{1}{c|}{SD}
        & \multicolumn{1}{c}{Observations}\\
        
        \midrule 
        1 & 60.61 & 0.87 & 3130 & 60.07 & 0.86 & 3217 & 61.21 & 0.81 & 3618\\
         
        2 & 57.96 & 0.89 & 3094 & 60.60 & 1.07 & 2086 & 61.99 & 1.13 & 1852\\
         
        3 & 58.52 & 0.89 & 3036 & 60.36 & 1.15 & 1825 & 61.44 & 1.20 & 1653\\
        \bottomrule
    \end{tabular}
    \caption{Empirical cumulative average reward (email open rate), standard deviation, and total allocation per arm, per policy, at the end of the real-world deployment.}
    \label{tab:empirical_data}
\end{table*}

Post experiment, we checked to what degree the average reward (mean open rate) differed across arms under each policy to determine whether we assigned to optimal conditions.  We used Wald \textit{z}-tests to compare if the arms' means have a statistically significant difference, taking into account their respective standard errors \cite{wald1943tests}. Since we have three arms and three policies, by fixing each policy we are able to compare pairs of arms under each. The results can be found in Table \ref{tab:wald_test_allocations}. To avoid making type I errors in this preliminary work, we controlled for a family-wise false positive rate of $0.05$ using a conservative Bonferroni Correction for every set of three—relevant—hypothesis test. 

Table \ref{tab:wald_test_allocations} also shows the \textit{p}-value for such tests, where we can see that all of them are greater than the adjusted-threshold of $0.017$, failing to reject the null hypothesis of all real means being equal. This result indicates within this adaptive policy allocation, it is unlikely that the arm means were different across all weeks.

\begin{table*}
    \centering
    \begin{tabular}{c|c|c|c|c|c|c}
        \toprule
        & \multicolumn{2}{c|}{UR}
        & \multicolumn{2}{c|}{TS}
        & \multicolumn{2}{c}{$\text{TS}^\dagger$}\\
        
        & \textit{p}-value
        & Wald statistic
        & \textit{p}-value
        & Wald statistic
        & \textit{p}-value
        & \multicolumn{1}{c}{Wald statistic}\\
        
        \midrule 
        Arm 1 vs. Arm 2 & 0.033 & 2.129 & 0.701 & 0.384 & 0.573 & 0.564\\
         
        Arm 2 vs. Arm 3 & 0.653 & 0.449 & 0.880 & 0.151 & 0.738 & 0.335\\
         
        Arm 3 vs. Arm 1 & 0.095 & 1.668 & 0.840 & 0.202 & 0.872 & 0.161\\
        \bottomrule
    \end{tabular}
    \caption{\textit{P}-values and statistics from the Wald \textit{z}-test by comparing the empirical cumulative arms' rewards within policies. Bonferroni adjusted threshold $p$=0.017.}
    \label{tab:wald_test_allocations}
\end{table*}

Finally, we conducted exploratory data analysis on the correlation between email open rate and their desired effect on student behaviour: earlier start times and higher homework attempt rate. We have not found any significance on their homework start times as of yet, but there is preliminary evidence that students who opened the emails have a higher attempt rate than those who did not.

%% file: sections/6_discussion.tex
\section{Discussion}




This work outlines a deployment framework for using adaptable multi-armed bandits for randomized testing. The goal was to maintain statistical power while giving students more optimal arm assignments. In our particular setting, multiple factors could have affected students' email engagement, such as midterm tests, coursework deadlines, etc. Moreover, student patterns may change across the semester, motivating us to explore ways of using an adaptive algorithm that has existing information about the probability of each arm being optimal, while getting updated current probabilities from the uniform data. We then formed the $\text{TS}^\dagger$ algorithm that learns from both the traditional TS and UR algorithms.

From our results, even though Table \ref{tab:wald_test_allocations} shows no statistically significant difference between arm means under all policies, there is an empirical mean increase of $4.6\%$ and $3.6\%$ from arm 1 compared to arms 2 and 3. This empirical difference causes TS and $\text{TS}^\dagger$ to extremely favour assigning to one arm, which only begins to change after multiple weeks. Their respective plots in Figure 1 illustrate this behaviour. These two points combined highlight the problem that MAB algorithms encounter: choosing an arm even when neither is more optimal than the other based only on average rewards. In other words, in the situation arm means are not significantly different, uniform random would be preferable as it provides more balanced data without erroneously determining an optimal arm to exploit.

Interestingly, Figure 1 shows that $\text{TS}^\dagger$ was able to hold on to the higher allocation of arm 1 for much longer than TS. However, this observation is counterintuitive, since our goal was to maintain more uniformity of allocations when conditions become equal. This further demonstrates that $\text{TS}^\dagger$ still suffers from the lack of this uniform guarantee, similar to regular TS, and instead seeks to allocate an arm rapidly, even without certainty of its optimality. Furthermore, one point of tension for adaptive experimentation is to know when to switch or adapt to a different allocation rather than hold on to a previously highly allocated condition which may no longer be optimal. In this environment, it seems that the TS algorithm can ``adapt" better than $\text{TS}^\dagger$, at the risk of greater uncertainty of allocation optimality.



Furthermore, sample estimates in Figure 1 hint at a non-stationary behaviour across weeks, meaning the optimal arm could change each week. This is a characteristic that can cause unexpected behaviour in adaptive algorithms to perform as expected, as bandit allocations such as TS assume constant arm means.

Our results highlight a lesson for researchers interested in deploying adaptive interventions in education: one must be certain of their student behaviour first. In the case that conditions are not significantly different, these algorithms lose statistical power compared to uniform random (i.e. traditional A/B testing) by favouring one arm over others. This can hinder instructors' ability to make beneficial choices to students based on experimental results.

In our case, we were not assigning students to a significantly poorer condition and the experiment had a simple manipulation. However, this was only able to be determined later in the experiment, not during the early deployments. In other settings with a higher risk for harm, for instance, most students could be assigned to an initially optimal arm—based on empirical reward—that turns out to be the least optimal arm in the long run. In a field such as computer science where topics build on top of previous lessons, it is important we only compound positive effects for students in our interventions. 




\section{Limitations and Future Work}
As discussed, the convergence for adaptive methods is still too rapid, even for non-significant differences in a dynamic environment. This could indicate that traditional uniform algorithms could be useful in controlling the number of students assigned to an optimal condition in the face of changing optimal arms each week. However, different analyses should be made to test for other possibilities. For example, it could be that the subject lines were too general to cause a behavior change, or even too effective in doing so. It could also be related to having the email itself be effective, regardless of the subject line. 

Furthermore, we recognize that unobserved factors in our study that could influence students' decision to open or not open an email, beyond the subject line. For instance, the engagement of a student with a course could vary throughout a semester. Also, this study was made on a single CS1 course. Thus, it may not be representative across domains.

Finally, our study encounters the problem of personalization, as the updates are made in a weekly batch to ensure time does not affect the decision of students' to open an email. However, this comes at the cost of the individualization of adaptive experimentation and does not allow for a finer-grained feedback structure beyond the weekly subject line.

Future work will focus on different approaches to addressing the problem of potentially non-stationary arm means. Even more, there is room to focus on adaptive frameworks that consider contextual factors to better predict student behaviour during, and not after, an experiment. Lastly, an additional investigation could design a personalized variant of bandits for email reminders that do not depend on a large batch size or reward collected from peers. 


%% file: sections/7_conclusion.tex
\section{Conclusion}


We present and evaluate a deployment of email reminders in a first-year computer science course which leverages Multi-Armed Bandits adaptive algorithms. The goal was to increase student engagement with our messages by manipulating subject lines relative to a uniform allocation. On a high level, these reminders are aimed at guiding computer science students to build the time management skills necessary for success in the field. Our work highlights the results from using adaptive algorithms in a real-world setting, and explains problems we encounter such as assigning higher allocation probability to an arm when there is no significant different among the arms' means. We recognize there could be outside factors that can affect the reward every iteration. Additional work should be done to address these and the evidence of non-stationarity in more depth. 

\section{Acknowledgements}
This work was partially supported by the Natural Sciences and Engineering Research Council of Canada (NSERC) (No. RGPIN-2019-06968), as well as by the Office of Naval Research (ONR) (No. N00014-21-1-2576)